%% file: colm2025_conference.tex
\documentclass{article} 
\usepackage[T1]{fontenc}
\usepackage[preprint]{colm2025_conference}

 \usepackage{multirow}
\usepackage{microtype}
\usepackage{hyperref}
\usepackage{url}
\usepackage{booktabs}
\usepackage{wrapfig}
\usepackage{graphicx}
\usepackage{caption}
\usepackage{amsmath} 
\usepackage{subcaption}
\usepackage{comment}
\usepackage{tabularx} 
\usepackage{longtable}

\usepackage{lineno}

\definecolor{darkblue}{rgb}{0, 0, 0.5}
\hypersetup{colorlinks=true, citecolor=darkblue, linkcolor=darkblue, urlcolor=darkblue}

\makeatletter
\renewcommand{\sectionautorefname}{\S\@gobble}
\renewcommand{\sectionautorefname}{\S\@gobble}
\renewcommand{\subsectionautorefname}{\S\@gobble}
\renewcommand{\sectionautorefname}{\S\@gobble}
\renewcommand{\subsectionautorefname}{\S\@gobble}
 \renewcommand{\appendixautorefname}{\S\@gobble}
\makeatother

\definecolor{purp}{HTML}{791f87}
\definecolor{highlight}{RGB}{255, 255, 0}
\definecolor{bottlegreen}{rgb}{0.0,0.42,0.31}
\definecolor{bblue}{rgb}{0.0, 0.6, 0.8}

\newcommand{\benchmark}{\textsc{BLAB}}
\newcommand{\benchmarkmini}{\textsc{BLAB-mini}}

\title{BLAB: Brutally Long Audio Bench}

\author{\ \ \ \ \ \ \ \ \ \ Orevaoghene Ahia\textsuperscript{1} \ \ Martijn Bartelds\textsuperscript{2} \ \ Kabir Ahuja\textsuperscript{1}  \ \ Hila Gonen\textsuperscript{1} \\ \ \  
\textbf{Valentin Hofmann\textsuperscript{1}} \ \ \textbf{Siddhant Arora \textsuperscript{4}} \ \ \textbf{Shuyue Stella Li\textsuperscript{1}} \ \ \textbf{Vishal Puttagunta\textsuperscript{3}} \\ \ \  \textbf{ Mofetoluwa Adeyemi\textsuperscript{5}} \ \ \textbf{Charishma Buchireddy\textsuperscript{3}} \ \ \textbf{Ben Walls\textsuperscript{3}} \ \ \textbf{Noah Bennett\textsuperscript{3}} \\ 
\ \ \textbf{Shinji Watanabe\textsuperscript{4} \ \ Noah A. Smith\textsuperscript{1} \ \ Yulia Tsvetkov\textsuperscript{1} \ \ Sachin Kumar\textsuperscript{3}} \\ \\ \ \ \  \ \ 
\textsuperscript{1}University of Washington \ \ \textsuperscript{2}Stanford University \ \  \textsuperscript{3}The Ohio State University \ \  \\ \ \  \ \ \ \ \  \ \ \ \  \ \ \ \  \ \ \ \ \textsuperscript{4}Carnegie Mellon University \ \ \textsuperscript{5}University of Waterloo  \\
\ \ \  \ \ \ \ \ \ \ \ \  \ \ \  \ \ \  \ \ \  \ \ \ \ \ \ \ \ \ \ \ \ \ \ \ \ \ \ \ \ \ \ \ \ \ \ \ \ \ \  \href{mailto:oahia@cs.washington.edu}{\texttt{oahia@cs.washington.edu}}
}

\begin{document}

\ifcolmsubmission
\linenumbers
\fi

\maketitle

\begin{abstract}
Developing large audio language models (LMs) capable of understanding diverse spoken interactions is essential for accommodating the multimodal nature of human communication and can increase the accessibility of language technologies across different user populations. Recent work on audio LMs has primarily evaluated their performance on short audio segments, typically under 30 seconds, with limited exploration of long-form conversational speech segments that more closely reflect natural user interactions with these models.
We introduce Brutally Long Audio Bench (\benchmark{}), a challenging long-form audio benchmark that evaluates audio LMs on localization, duration estimation, emotion and counting tasks using audio segments averaging 51 minutes in length. 
\benchmark{} consists of 833+ hours of diverse, full-length audio clips, each paired with human-annotated, text-based natural language questions and answers. Our audio data were collected from permissively licensed sources and underwent a human-assisted filtering process to ensure task compliance. We evaluate six open-source and proprietary audio LMs on \benchmark{}, and find that all of them, including advanced models such as Gemini 2.0 Pro and GPT-4o, struggle with the tasks in \benchmark{}. Our comprehensive analysis reveals key insights into the trade-offs between task difficulty and audio duration. In general, we find that audio LMs struggle with long-form speech, with performance declining as duration increases. They perform poorly on localization, temporal reasoning, counting, and struggle to understand non-phonemic information, relying more on prompts than audio content. 
\benchmark{} serves as a challenging evaluation framework to develop audio LMs with robust long-form audio understanding capabilities. \footnote{We provide data and code at \url{https://github.com/orevaahia/brutally_long_audio_bench}}

\end{abstract}

\input{01}

\input{02}
\input{03}

\input{04}
\input{05}

\input{06}

\section*{Ethics Statement}
\benchmark{} is entirely sourced from Creative Commons-licensed videos on YouTube, using a rigorous human-assisted filtering process to ensure diverse and high-quality content. Human speech is a particularly sensitive modality, as it is deeply personal and can convey not only language but also emotions and mental states. Each audio file in \benchmark{} is carefully selected, with deeply sensitive material excluded to protect privacy. We have also ensured that no child sexual abuse material is present in our dataset.\footnote{\url{https://www.missingkids.org/cybertiplinedata}}
We recognize that speech recordings can be used to track or identify individuals without their knowledge or consent. To address this, we have built our dataset using publicly available YouTube data that complies with ethical guidelines regarding privacy and data usage. However, we also acknowledge the potential risks of data misuse, such as the unintended identification of individuals or the reinforcement of biases in downstream audio language models due to potential contamination. Therefore, we encourage responsible use of our dataset and highlight the importance of considering privacy and ethical concerns when applying it to evaluate real-world applications. To promote transparency and reproducibility, we will make our benchmark publicly available, along with associated evaluation metrics and data curation framework, allowing the research community to contribute and build upon our work.

\section*{Acknowledgments}
We thank the UW NLP community for valuable discussions on this work. We are grateful to Inna Lin, Melanie Sclar, Kyle Lo, Lucy Lin, Vidisha Balachandran for discussions on data curation, experiments, and analysis. This work was supported in part by the \href{https://www.osc.edu/}{Ohio Supercomputer Center}.
This research was developed in part with funding from the Defense Advanced Research Projects Agency's (DARPA) SciFy program (Agreement No. HR00112520300), and NSF grant No.~IIS2203097. The views expressed are those of the author and do not reflect the official policy or position of the Department of Defense or the U.S.~Government.
We also gratefully acknowledge gift funding from Google, MSR, and OpenAI.



\bibliography{colm2025_conference}
\bibliographystyle{colm2025_conference}

\appendix
\input{appendix}


\end{document}

%% file: 01.tex
\section{Introduction}
\begin{figure}[t!]
  \centering
  \resizebox{\textwidth}{!}{
  \includegraphics[width=\linewidth]{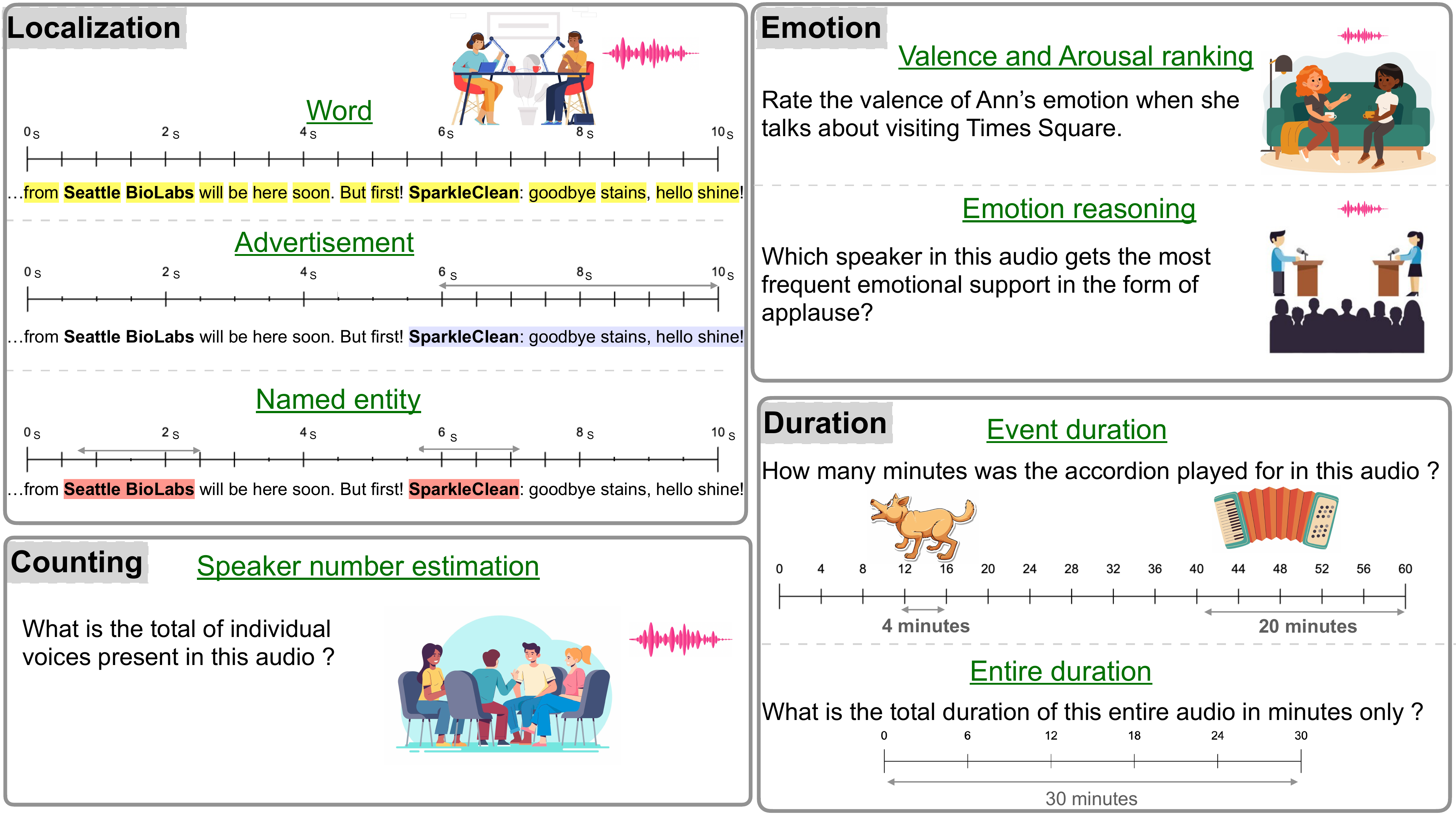}}
    \caption{Overview of \benchmark{}, designed to test true multimodal understanding abilities of audio LMs. It contains eight distinct audio tasks across four categories, namely \textbf{localization}, \textbf{counting}, \textbf{emotion}, and \textbf{duration estimation}. $^\dag$\scriptsize All images are designed by \href{https://www.freepik.com/}Freepik .}
  \label{fig:benchmark}
  
\end{figure}

Developing large audio language models (LMs) is essential for increasing accessibility of language technologies~\citep{Qwen2-Audio, geminiteam2024gemini15unlockingmultimodal, ghosh-etal-2024-gama, openai2024gpt4technicalreport, tang2024salmonn, ghosh2025audioflamingo2audiolanguage,microsoft2025phi4minitechnicalreportcompact}.
Text alone cannot fully capture the richness of human communication, which is inherently multimodal, including not just words but a wide range of auditory cues, such as tone, pitch, and rhythm.
However, we lack a comprehensive understanding of current audio LM capabilities, especially for longer audio fragments of conversational speech.
Existing benchmarks consider isolated utterances or short audio clips up to 30 seconds~\citep{huang2024dynamicsuperbphase2collaborativelyexpanding, sakshi2024mmaumassivemultitaskaudio, yang-etal-2024-air} or at most 5 minutes \citep{ghosh2025audioflamingo2audiolanguage},
failing to account for challenges associated with long-term dependencies \citep{Chen2024TrainLA}, temporal relationships, context retention, and evolving prosodic cues in long audio durations.
The temporal dimension is particularly important for understanding long-form content, where events unfold sequentially rather than as isolated units.

To address this gap, we introduce Brutally Long Audio Bench (\benchmark{}), a comprehensive benchmark for understanding and reasoning over audio samples with durations between 15 minutes and 2 hours. \benchmark{} is comprised of 833+ hours 
of conversational speech content
across eight challenging tasks under four skill categories underlying fundamental tasks, namely: localization, counting, emotion, and duration estimation.
For instance, \benchmark{} assesses an audio LM's ability to predict timestamps (localization) necessary for audio retrieval and event detection, which, to the best of our knowledge, is not evaluated in existing benchmarks. Compared to existing benchmarks, \benchmark{} is substantially more challenging due to the extensive length of the audio samples, which contain richer contextual information. Our data is entirely sourced from Creative Commons-licensed videos on YouTube, using a rigorous human-assisted filtering procedure to ensure diverse and high-quality content (more details in \autoref{sec:methodology}).

Using \benchmark{}, we conduct a comprehensive evaluation (\autoref{sec:results}) and in-depth analysis (\autoref{sec:analysis}) of several frontier audio LMs.
Our analysis reveals that even proprietary models, achieve an average $F_1$ score up to  \textbf{3.02} on localization tasks (Gemini 2.0 Flash) and average exact match accuracy up to \textbf{22.25} on the remaining tasks (Gemini 2.0 Pro), underscoring the complexity of our benchmark and the limitations of current modeling approaches. 
We thoroughly analyze model responses across all tasks and document key patterns, common errors, and areas where models struggle the most. We find that audio duration plays a large role in model performance in \benchmark{}, as well as task complexity. Even though models struggle to perform tasks in \benchmark{}, we still observe considerable performance gaps between open-sourced and proprietary models, especially Gemini.  

These findings motivate new research on long-form audio processing. But the lack of transparency regarding the training data and checkpoints of most models makes it challenging to thoroughly probe their results and understand their underlying mechanisms. This highlights the urgent need for developing long-context open-source multimodal language models trained on audio and text, where  the data, model checkpoints, and training techniques are fully documented and accessible. We will release all our data, code and model generations for reproducibility and further research by the community.

%% file: 02.tex
\section{BLAB: Brutally Long Audio Bench}
\label{sec:methodology}
\benchmark{} evaluates audio LMs that accept both text and audio as input and generate text.
The primary focus of the benchmark is to evaluate audio perception and reasoning abilities of models on long-form audio recordings derived from various real-world sources, such as interviews, podcasts, and political speeches.  \benchmark{} includes eight distinct tasks across four categories, namely localization, counting, emotion, and duration estimation.
Each task includes 200 Creative Commons-licensed audio files sourced from YouTube, alongside corresponding human-annotated, text-based questions and answers.
Each audio sample was carefully and thoughtfully selected, prioritizing complexity, quality, diversity, and task relevance. The question-answer pairs were either manually generated by the authors of this paper
or model-generated and verified by the authors to ensure quality.
We designed each task to test true multimodal understanding abilities of audio LMs.
The tasks in BLAB are likely difficult to solve with cascaded systems that first transcribe the audio and then query a text-only LM, as they rely on information in the audio signal that is typically lost during the transcription process. We provide a detailed description of each category and its tasks below.

\begin{table}[t!]
\centering
\resizebox{1.0\textwidth}{!}{
        \begin{tabular}{p{3cm} rr p{11cm}} 
            \toprule
            \textbf{Task} & \textbf{TD (hrs.)} & \textbf{AD (mins.)} & \textbf{Question Example} \\
            \midrule
           Word Localization  & 191 &  57 & Align the provided audio file with each word in its transcript. For each word spoken, predict the start and end timestamps in seconds and milliseconds. \\
           \midrule
           NE Localization  & 110 & 56 & Your task is to analyze an audio file and detect all \textbf{Movie} named entities present within it. \textbf{Movie} entity refers to the name of \textbf{feature films or animated movies}. For each detected \textbf{Movie} entity, provide the \textbf{start} and \textbf{end} timestamps (in seconds and milliseconds) that indicate the location and duration of the entity in the audio. \\
           \midrule
            Advertisement Localization & 232 & 70 & Given the audio file, your task is to detect the presence of promotions and advertisements within the audio. You are required to predict the start and end timestamps (in seconds and milliseconds) for the entire span for all promotion and advertisement segments you have identified. \\
            \midrule
           Speaker Number Estimation  & 176 & 53 & What's the number of unique voices detected in this sound file? \\
           \midrule
           Valence \& Arousal Ranking & 18 & 39 & How would you assess the valence of the speaker in the fourth apology clip within this audio file? \\
            \midrule
           Emotion Reasoning & 16  & 44 &  Throughout the course of this debate, which speaker demonstrates the most consistent emotional intensity, particularly in their tone and delivery? \\
           \midrule
            Entire Duration & 92 & 27 & What is the total duration of this audio file in seconds only? \\
            \midrule
           Event Duration & 174 & 51 & A harmonica is being played fifteen seconds towards the end of this audio. How long in seconds was this instrument played for ? \\
            \bottomrule
        \end{tabular}}
    
    \caption{Statistics and examples of questions for each task in \benchmark{}. TD and AD represent total and average duration, respectively. Each task consists of 200 question-audio-answers.}
    \label{tab:dataset_stats}
\end{table}

\subsection{Localization Tasks} 

The localization tasks require identifying the temporal positions of specific events with varying durations within audio samples by outputting the precise start and end timestamps at which those events occur.
Localization tasks are relevant in real-world applications, such as audio retrieval and event detection, where precise temporal alignment is important to achieve user goals.
Despite their importance, such tasks have been absent from existing benchmarks or limited to 30-second audio inputs \citep{huang2024dynamicsuperbphase2collaborativelyexpanding, sakshi2024mmaumassivemultitaskaudio, yang-etal-2024-air}.
Within this category, we create three tasks, namely \textbf{word localization}, \textbf{advertisement localization}, and \textbf{named entity (NE) localization}.  In each case, the target output is one or more pairs of start/end timestamps.
We describe our annotation procedure below.

\paragraph{Word Localization} We used 200 audio files obtained from YouTube (total duration of 191 hours, samples are 57 minutes on average) and applied existing forced alignment (FA) tools to obtain word- and sentence-level alignments between audio samples and their corresponding transcripts.
For word localization, we used WhisperX to generate word-level timestamps for each word spoken in each audio sample~\citep{bain2022whisperx}.
Next, an annotator (one of the authors) manually reviewed a subset of our entire dataset to ensure that the forced alignments were error-free (only $\sim$1\% of timestamps needed to be corrected). Each audio sample contains 10,500 word-timestamp pairs on average.

\paragraph{Named Entity Localization}
We defined nine broad entity categories to be localized: Event, Location, NORP (nationalities or religious or political groups), Organization, Person, TV shows, Temporal, and Work of Art, as well as ``All entities'', which includes all of the above. For each broad category, we also define fine-grained subcategories, allowing us to evaluate whether a model understands the nuances of entity types within the same category. For instance, the example in \autoref{tab:dataset_stats} focuses on movie entities, which are a subcategory of Work of Art.
Next, we crawled 200 audio files from YouTube and transcribed the audio using WhisperX. Each transcript was then fed into a text-only language model to extract plausible NE spans for all entity categories. We tested this part of the annotation process with GPT-4~\citep{openai2024gpt4technicalreport} and Claude-3~\citep{TheC3} and obtained a higher recall with Claude-3 so we settled on Claude-3.
After extracting the NE spans from text, we mapped these spans back to their timestamps (their location in the audio), also derived from WhisperX. We paired 49 audio files from our pool to the ``All entities'' category. For the single entity categories, we excluded audio files with fewer than 15 predicted entity spans or a duration of less than 20 minutes, leaving 69 audio files from the remaining 151 in our pool. These 69 audios were then paired to multiple ``single entity'' categories, resulting in 151 extra items for the NE localization task.   
This yields audio data with a total duration of 110 hours, each sample being 56 minutes on average.
Compared to existing work by \citet{huang2024dynamicsuperbphase2collaborativelyexpanding}, which reports an average of 2 entities per sample, our dataset contains an average of 46 entities per sample. The number of entities to be localized and their temporal position in the audio adds to the complexity of this task, as models often struggle to capture long-term dependencies.

\paragraph{Advertisement Localization} We used 200
podcasts from YouTube using the ``hasPaidProductPlacement'' filter in the YouTube API. Similar to the NE localization annotation procedure, we transcribed each podcast using WhisperX.  We  fed the transcripts into GPT-4 to extract plausible advertisement segments. Since we obtained very long transcripts, we fed them to the model in chunks of 20 sentences at a time, instructing the model to extract segments that contained an advertisement of a product or promotion from the podcast host.
It appeared that GPT-4 sometimes incorrectly identified segments of the transcript as advertisements, while in other cases it missed spans that should have been detected. To remedy this, an annotator (one of the authors) reviewed all predicted advertisement segments for every audio sample, removed false positives and added segments that were missing. The percentage of advertisement segments that were kept was 84\%. Subsequently, we aligned human-verified advertisement segments with their temporal location in the audio file using the sentence-level timestamps from WhisperX. Our final set of audio data consists of 232 hours of audio across 200 podcasts, with an average of three advertisement segments per podcast and 180 words per advertisement segment.

\subsection{Counting Task}
\paragraph{Speaker Number Estimation} The task of this category is to count the total number of distinct voices in an audio sample. 
Existing benchmarks only feature audio samples with fewer than four distinct speakers \citep{huang2024dynamicsuperbphase2collaborativelyexpanding, sakshi2024mmaumassivemultitaskaudio, yang-etal-2024-air}. In contrast, \benchmark{} includes audio samples with speaker counts ranging from four to 80. This introduces several challenges. First, such long audio samples often contain overlapping speech, making the task particularly difficult for models that struggle with speaker diarization, i.e., attributing speech segments to individual speakers.
Second, the samples are 53 minutes long on average,
with new speakers appearing later in the audio samples. This poses difficulties for models that struggle to process long speech.

We included 200 YouTube audio samples (total duration of 177 hours with an average audio sample length of 53 minutes) in this task. Two annotators (also authors) listened to the entire audio samples and counted the number of distinct speakers. Due to the complexity of the task, some audio samples received different counts from the annotators. For the cases with different counts, a third annotator reviewed both annotations and selected a final count label. In the majority of cases, annotators argued that multiple counts could be valid, so we retained a range of count labels and adjusted our evaluation metric (see Section~\ref{sec:metrics}) to consider any number within the range as correct. Overall, audio samples with a range of values as their ground truth count labels account for 60\% of our dataset, and the range does not exceed two speakers in general.

\subsection{Emotion Tasks} 
The emotion tasks involve ranking emotions expressed in speech and non-verbal sounds, as well as reasoning over emotional expressions in audio.
They evaluate a model's ability to integrate  semantic content with acoustic features that are strong indicators of emotions, and are not available in text-only representations.

\paragraph{Valence \& Arousal Ranking} 
Existing benchmarks typically structure emotion tasks around discrete emotion classification, using labels such as happy, sad, or angry~\citep{huang2024dynamicsuperbphase2collaborativelyexpanding, sakshi2024mmaumassivemultitaskaudio, yang-etal-2024-air}. However, this design does not account for variations in emotional intensity, which is particularly relevant in speech-based emotional expressions \citep{6883166, Sethu2019TheAW}. This motivates us to evaluate audio LMs' ability to rank ordinal emotional expressions in audio segments rather than to classify them. \citet{Yannakakis_2017, Yannakakis_2021} presented strong evidence supporting the ordinal nature of emotions, demonstrating that ordinal labels enhance the validity, reliability, and overall performance of emotion recognition models in affective computing. Ordinal emotion labels from classical emotion theory \citep{Russell1980, Lotfian_2019_3} are based on ranking emotions by intensity, and are often analyzed along three dimensions: valence (the degree of negativity or positivity in the emotion), arousal (level of activation or intensity), and dominance (control or power over the emotion). 

To curate our dataset, we used 28 audio samples obtained from YouTube (total duration is 18 hours, and samples are 39 minutes on average) and performed annotation through human-AI collaboration. To increase the complexity of the task, we focused on ranking speech segments in an audio file that are difficult to distinguish in terms of emotional content based off an audio transcript. This challenges cascaded systems and encourages end-to-end audio LMs to jointly consider the prosodic and semantic content in the audio. Each audio sample in our dataset is diarized into speaker-specific segments using Pyannote~\citep{Plaquet23, Bredin23}, obtaining 600 segments on average per audio sample. To create instances for which labels cannot be inferred from text alone, we input each audio segment into an emotion recognition model trained to predict valence and arousal scores~\citep{wagner2023dawn}. Next, we input the corresponding transcripts to GPT-4 to obtain text-based valence and arousal scores. We filter segments for which the difference between the audio and text-based valence and arousal scores for the same segment was greater than a threshold (0.3 in our experiments). Two annotators reviewed each filtered segment, verifying its alignment with the expressed emotion. For each sample, we then randomly sample up to four segments for evaluating both valence and arousal. We then crafted specific questions for each segment, prompting the model to rank the expressed emotion. This resulted in 156 high-quality segments with verified valence and arousal scores across 18 audio samples. 

\paragraph{Emotional Reasoning}
Our motivation for this task is to evaluate an audio LM's ability to understand emotions beyond surface-level sentiments in long audio, focusing on deeper emotional states, contextual cues, and pragmatic meanings of speech over extended periods. We manually identified 22 audio samples on YouTube that were suitable candidates for this task, such as those obtained from controversial debate podcasts, comedy shows, movie review podcasts, and emotionally charged interviews. One annotator listened to each audio sample and identified  emotional patterns and shifts, such as transitions from calmness to agitation, or from happiness to frustration, and crafted reasoning questions based on verbal and non-phonemic cues, like changes in speech tempo, pauses, or arousal. We design this task as a multiple-choice question answering problem, where confounders are generated by extracting plausible, contextually relevant answers from the audio, increasing the task's difficulty. Overall we have 44 questions paired with 22 audio samples and plan to scale further in future work. 

\subsection{Duration Tasks}
\paragraph{Entire Duration}
In this task, an audio LM is tasked to predict the total duration of an audio file, expressed in seconds. The model should output a numerical value representing the duration of the audio sample. To create the dataset, we used 200 YouTube audio samples, which have a duration ranging between 8 seconds and 92 minutes. This is the only task in our benchmark that also contains short audio files.  
Ground truth labels are generated by converting the duration of each audio to seconds. The total duration of audio samples is 92 hours, and samples are 27 minutes on average. 

\paragraph{Event Duration}
This task involves predicting the duration of specific acoustic events (e.g., laughter in a comedy special, question-and-answer segments in a panel session, or a particular speaker's total speaking time in a meeting) within an audio sample, or the total duration of the sample itself in seconds. This task evaluates basic temporal reasoning abilities of audio LMs, including their capacity to identify, localize, and track both verbal and non-verbal acoustic events and activities. We used 87 YouTube audio samples (total duration of 77 hours with an average duration of 53 minutes). An annotator (one of the authors) reviewed each audio sample, detecting and labeling acoustic events and activities. The diverse range of genres in YouTube allowed us to identify multiple events within a single audio sample. After selecting these events, the annotator formulated questions and answers and added their specific durations, pairing 200 questions to all 87 audio files. Each audio sample was paired with at least one question and up to a maximum of four questions.

%% file: 03.tex
\section{Experimental Setup}
\subsection{Models}\label{sec:models}
Almost all open-weight audio LMs  \citep{10389742, Qwen2-Audio, ghosh2024gamalargeaudiolanguagemodel, tang2024salmonn}
can only process audio samples with a duration of 30 seconds or less.
This limitation comes from their underlying training setup
\citep{10.5555/3495724.3496768, radford2022robustspeechrecognitionlargescale},
which truncates audio samples longer than 30 seconds.
Audio Flamingo \citep{ghosh2025audioflamingo2audiolanguage} is an exception, as it supports audio samples with a duration of up to 5 minutes.
In contrast, closed-source Gemini models \citep{geminiteam2024gemini15unlockingmultimodal} support up to 9.5 hours of audio, while GPT-4o
\citep{openai2024gpt4technicalreport} handles up to 8 minutes of audio.
Based on these model characteristics and the design of \benchmark{}, our current evaluations are performed using only Gemini models, specifically Gemini 2.0 Flash and Gemini 2.0 Pro.
However, to ensure broader comparisons and for detailed analysis on the influence of audio duration on audio LMs performance, we evaluate other models \citep{Qwen2-Audio, openai2024gpt4technicalreport, ghosh2025audioflamingo2audiolanguage, microsoft2025phi4minitechnicalreportcompact} on a curated short audio ($\leq30$ seconds) subset of our benchmark named \benchmarkmini{}.

\subsection{Evaluation Strategy}
\paragraph{Setup}
The localization, counting, and entire duration tasks include 200 audio samples each, for which we handcrafted a set of 20 unique natural language questions per task. These questions are paraphrases of each other. For each audio sample, we uniformly randomly pair it with one of the questions from the corresponding task's question set. Thus, each question is paired with $\sim10$ audio samples, ensuring a diverse distribution of questions throughout the dataset.
For event duration and emotion tasks, each question is unique to an audio sample as they contain the event information present in the audio, resulting in 200 unique questions.

In our experiments, the audio LMs take a text prompt (instruction) and an audio file as input and produce text as output. In order to ensure that models truly understand the audio samples and avoid biases by providing options, we restrict our benchmark to free-form generation, except for emotion tasks (where the performance remains poor with or without options).
Also, prior work has suggested that multiple-choice question answering is not always reliable, as distractor options are often either too plausible or models exploit shortcuts to arrive at the correct answer \citep{palta-etal-2024-plausibly, balepur2025bestdescribesmultiplechoice}. For emotion, confounders are generated by extracting plausible, contextually relevant answers from the audio. We use greedy decoding for all models for reproducibility. 

\paragraph{Prompt Formatting} To ensure consistent outputs across different inputs and models, we append task-specific suffixes to the original questions. For localization tasks, these suffixes prompt the model to return JSON-formatted strings with start and end timestamps. For duration and counting tasks, models are instructed to output a number only, without additional explanation. In emotion tasks, which follow a multiple-choice format, the model is prompted to select the most appropriate option from the provided choices. We provide more details about our prompt formatting in \autoref{tab:prompt_format} in the Appendix.

\paragraph{Metrics} We use task-specific metrics to evaluate model performance on BLAB.
For all tasks, model outputs are post-processed to match the expected ground truth format.

For localization tasks, we compute precision and recall and both Frame-level $F_1$ (\textit{Named Entity Localization \& Advertisement Localization}), and Word-level (\textit{Word Localization}) $F_1$ scores. Predictions that are malformed such as those containing only a single timestamp, without indicating whether it is a start or end time are assigned a score of zero.
Counting, duration and emotion tasks are evaluated using exact match accuracy (EMA). For duration tasks, we also report EMA scores with an offset of $\pm$2 seconds to account for minor timing discrepancies.
More details about our evaluation metrics are given in Appendix~\ref{sec:metrics}.

%% file: 04.tex
\section{Results and Discussion}\label{sec:results}
In Table~\ref{tab:all_results}, we present the performance of Gemini 2.0 Flash and Gemini 2.0 Pro on \benchmark{}.
We discuss the performance on each of the tasks in more detail below.

\begin{table}[t]
    \centering
    \resizebox{\columnwidth}{!}{
    \begin{tabular}{cccc}
        \toprule
        \textbf{Task} & \textbf{Metric $(\uparrow)$ } & \textbf{Gemini 2.0 Flash} & \textbf{Gemini 2.0 Pro} \\
        \midrule
        Word Localization & word $F_1$  & 1.12 & 0.19 \\    
        \midrule
        Advertisement localization & Frame-level $F_1$  & 4.93 & 0.15  \\
         \midrule
        NE Localization & Frame-level $F_1$  & 2.97 & 2.14 \\
        \midrule
        Speaker Number Estimation  & EMA  & 8.00 & 8.50 \\
        \midrule  
        Valence and Arousal Ranking & EMA  & 26.28 & 32.00 \\ 
        \midrule
        Emotion Reasoning & EMA  & 54.54 & 64.29 \\
        \midrule
        Entire Duration & EMA (without / with  $\pm$2    seconds offset)   & 0.50/3.50   &  0.00/2.50 \\
        \midrule
        Event Duration & EMA (with / without $\pm$2   seconds offset) & 1.49/4.95& 1.49/3.96 \\

        \bottomrule
    \end{tabular}}
    \caption{Performance comparison of Gemini audio LMs across all \benchmark{} long audio tasks. Both models exhibit similar performance, generally achieving low performance across tasks.}
    \label{tab:all_results}
\end{table}

\subsection{Localization Tasks}
Word localization appears the most challenging task in \benchmark{} with both models performing extremely poorly. Both Gemini models achieved $F_1$ scores below 2\%.
These scores are particularly noteworthy, as state-of-the-art word timing models typically achieve scores close to 99\% on these last two metrics, as noted by \citet{Sainath2020EmittingWT}. \footnote{Out of 200 word localization examples, 26 predictions from Gemini Flash and 4 from Gemini Pro were malformed—often returning only a single value instead of the required start and end timestamps—and therefore received a score of zero.}We note that each audio sample for this task contains an average of $\sim$10200 words. Gemini, due to its limited output context length of 8096 tokens, is able to generate only $\sim$261 word timestamps per sample, accounting for only about 2\% of the ground truth. Gemini 2.0 flash achieves a precision score of 24.37 indicating that the model predictions are correct approximately 24.58\% of the time. However precision for Gemini pro is very low at 3.42. 

Models also perform poorly on NE and advertisement localization, with frame-$F_1$ scores below 5\%.
For NE localization, the Gemini models can detect 27\% of the ground truth entities, but fail to correctly locate them in the audio.
For advertisement localization, model performance is better when the advertisements are at the beginning of the audio files, and the Gemini models are more accurate at predicting start times than end times.
This leads us to hypothesize that the models estimate rather than detect segments with advertisements.

\begin{figure}[h]
  \centering

  \begin{subfigure}[b]{0.42\linewidth}
    \centering
    \includegraphics[width=\linewidth]{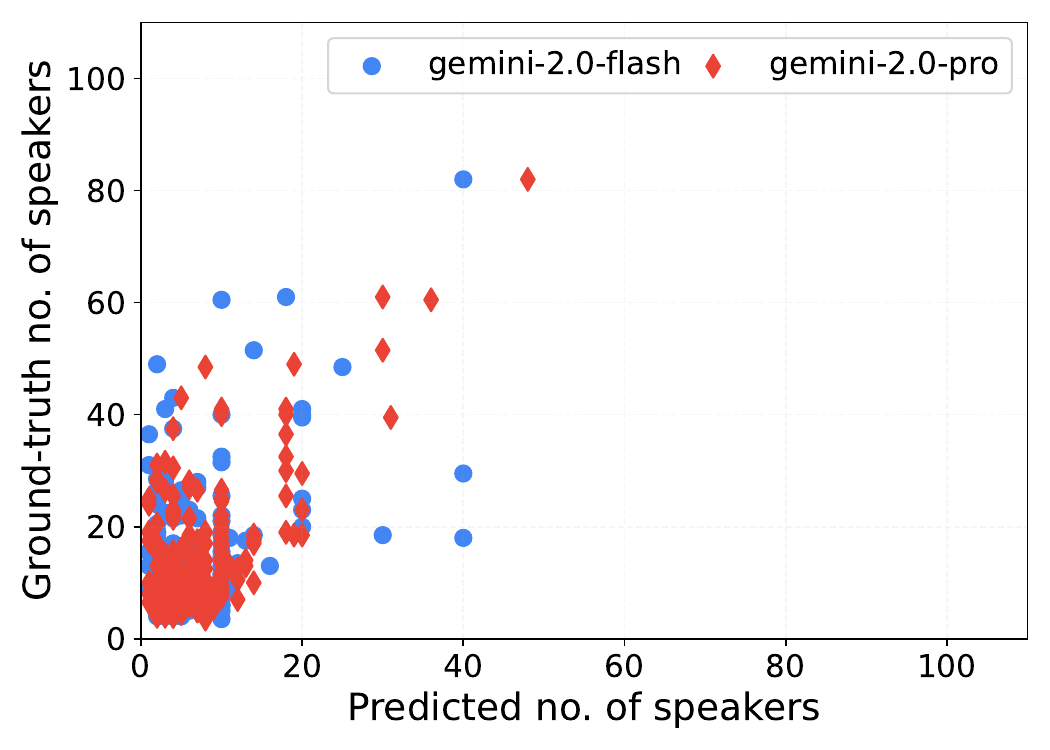}
    \caption{Gemini often underestimates the number of speakers in the speaker number estimation task}
    \label{fig:snv_analysis}
  \end{subfigure}
  \hfill
  \begin{subfigure}[b]{0.42\linewidth}
    \centering
    \includegraphics[width=\linewidth]{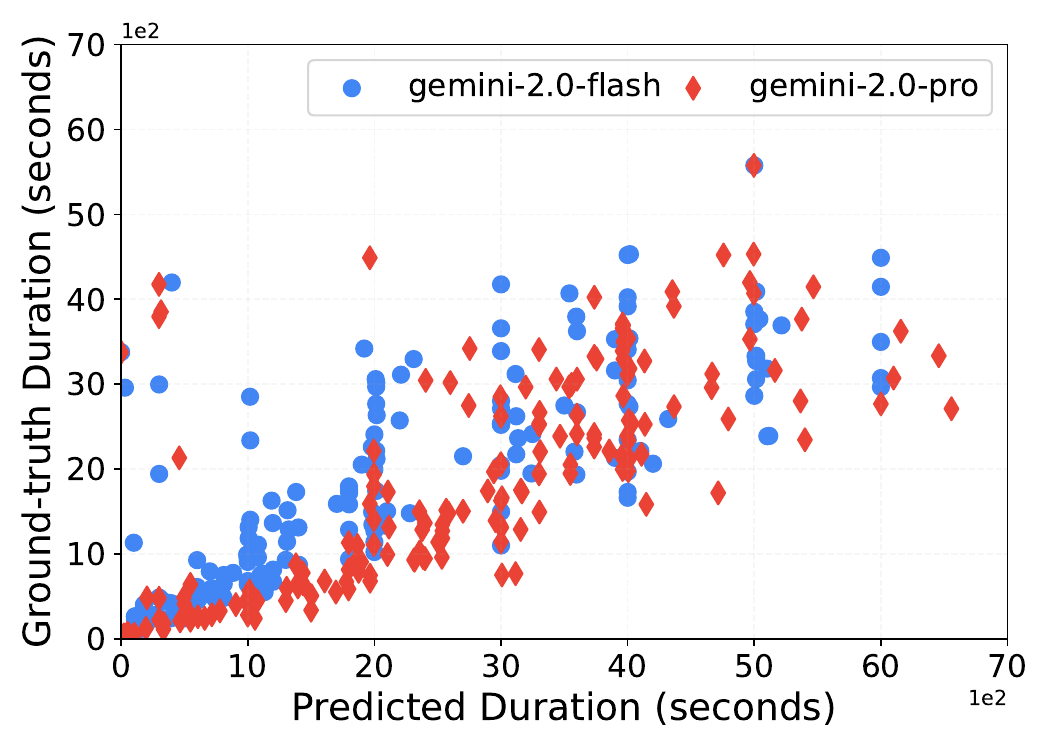}
    \caption{Gemini frequently overestimates the duration of audio samples in the entire duration task.}
    \label{fig:dur_analysis}
  \end{subfigure}
  \caption{Comparison of predicted and ground truth values for speaker number estimation and entire duration tasks}
  \label{fig:entire_duration_nd_snv_plots}
\end{figure}

\subsection{Counting Task}

The EMA on speaker number estimation for both models is below 9\%. 
Models typically underestimate the number of unique speakers (see Figure~\ref{fig:entire_duration_nd_snv_plots}a)
and are unable to distinguish between multiple speakers. We find that this error is exacerbated in instances with overlapping voices. 
In some cases, we observe overestimation, likely due to the models considering the same speaker at different positions in the audio as distinct.
These errors suggest that audio LMs lack the ability to track speakers consistently across turns, in conversations with overlapping speech, audio with music, commentary, other forms of extraneous content, or audio with varying prosodic features generally.

\subsection{Duration Tasks}
\paragraph{Entire Duration}
We find that Gemini struggles to predict the entire duration on full audio samples (EMA up to 3.50\%). Compared to our observations in the speaker number estimation task, our analysis indicates that the models often overestimate duration, as shown in \autoref{fig:entire_duration_nd_snv_plots}b. However, in most cases where the predictions are correct, the actual duration is less than 60 seconds.

\paragraph{Event Duration}
The performance scores are low as well for this task (EMA up to 4.95\%). From our observations, there are no clear trends regarding which acoustic events are predicted more accurately than others. Performance generally varies across different event types, and we observe that the model tends to underestimate event durations more frequently than it overestimates them. 

 \subsection{Emotion Tasks}
The best performance scores are obtained on the emotion-related tasks in \benchmark{}. Gemini 2 Pro outperforms Gemini 2 Flash on both tasks. Specifically, Gemini 2 Pro obtains an EMA of 32.00\% on the emotion ranking task, while it achieves an EMA of 63.63\% on emotion reasoning. For emotion ranking, particularly for arousal, we find that even Gemini 2 Pro struggles to correctly rank scenarios with extremely calm emotions, often misclassifying them as neutral or highly aroused. However, the model accurately predicts higher arousal levels in 80\% of cases. For valence, there are no observed trends, as accuracy across all rankings appears close to random chance.

%% file: 05.tex
\section{Analysis}\label{sec:analysis}
\paragraph{What role does the duration of audio play? }
In this section, we analyze the impact of audio duration on task performance by conducting experiments on shorter audio samples. These experiments also let us evaluate a broader set of LMs, in particular open-weights models that process audio inputs with a maximum of 30 seconds (Qwen 2, \citealp{Qwen2-Audio}, and Phi-4-Multimodal Instruct,
\citealp{ microsoft2025phi4minitechnicalreportcompact}) and 5 minutes (Audio Flamingo, \citealp{ghosh2025audioflamingo2audiolanguage}, and GPT-4o, \citealp{openai2024gpt4technicalreport}). We conduct our analyses on word and entity localization, speaker number estimation and both duration tasks.
We derive the data used in these experiments from samples in \benchmark{} by extracting audio segments up to 30 seconds. We reuse pre-existing annotations for the localization task and re-annotate the segments for speaker number estimation and duration tasks using the same procedure described in \autoref{sec:methodology}. We refer to this dataset as \benchmarkmini{}. In total, \benchmarkmini{} contains 813 questions and 346 minutes of audio in total. More details are provided in Appendix \autoref{tab:short_audio_data_statistics}.

\begin{figure}[ht]
  \centering
  \begin{subfigure}[b]{0.40\linewidth}
    \centering
    \includegraphics[width=\linewidth]{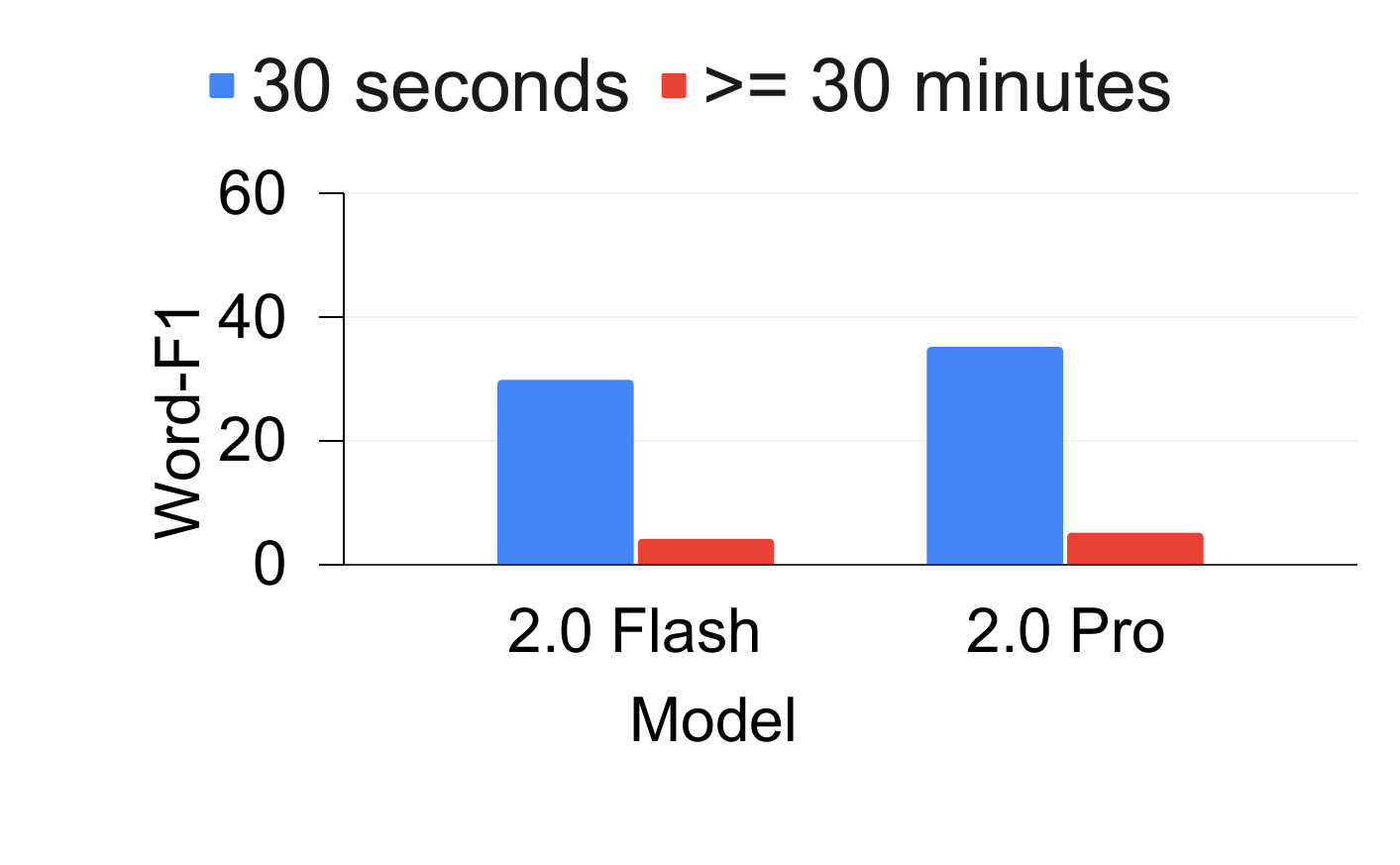}
    \caption{Word Localization}
    \label{fig7:a}
  \end{subfigure}
  \begin{subfigure}[b]{0.40\linewidth}
    \centering
    \includegraphics[width=\linewidth]{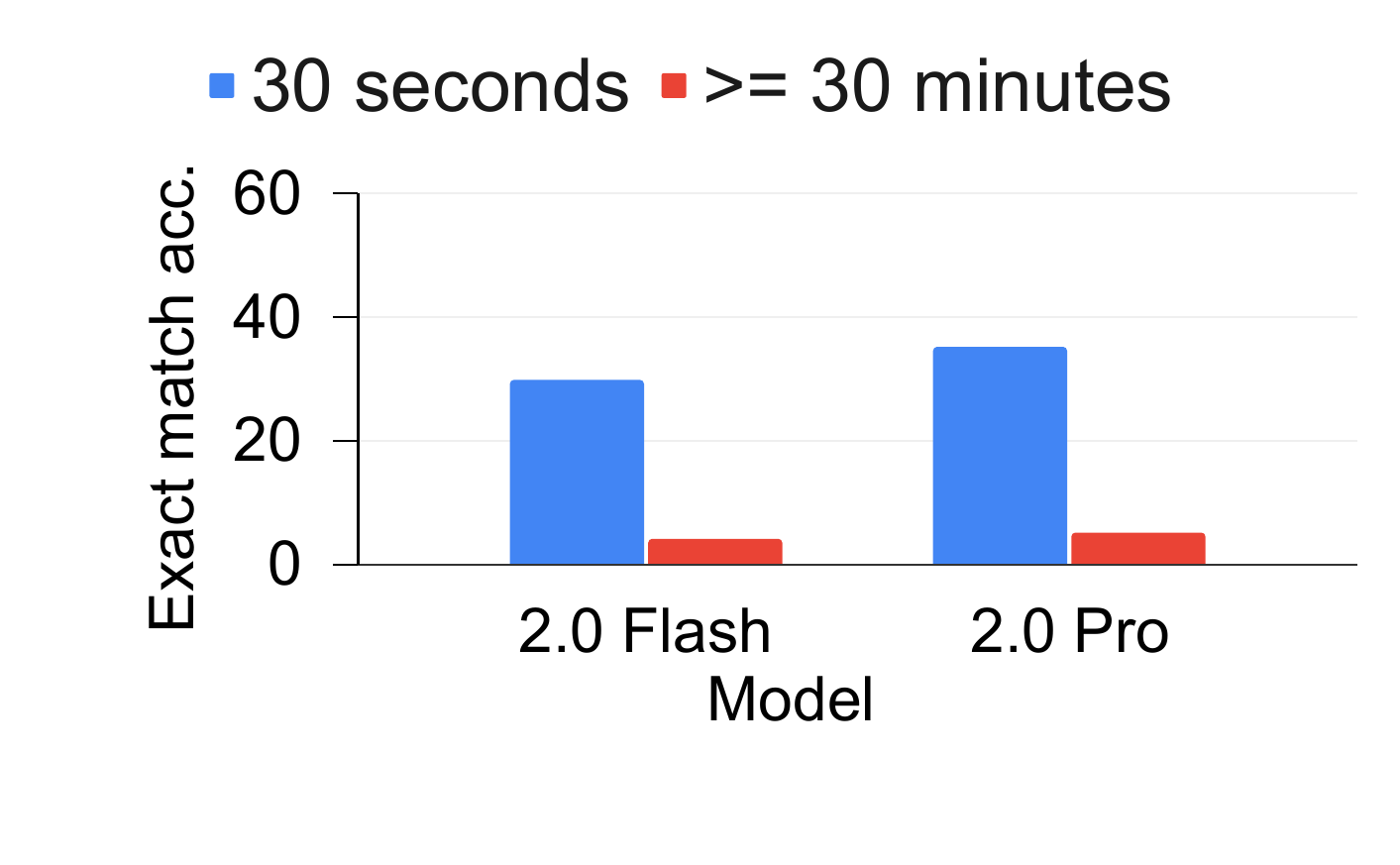}
    \caption{Event Duration}
    \label{fig7:b}
  \end{subfigure}
  \begin{subfigure}[b]{0.40\linewidth}
    \centering
    \includegraphics[width=\linewidth]{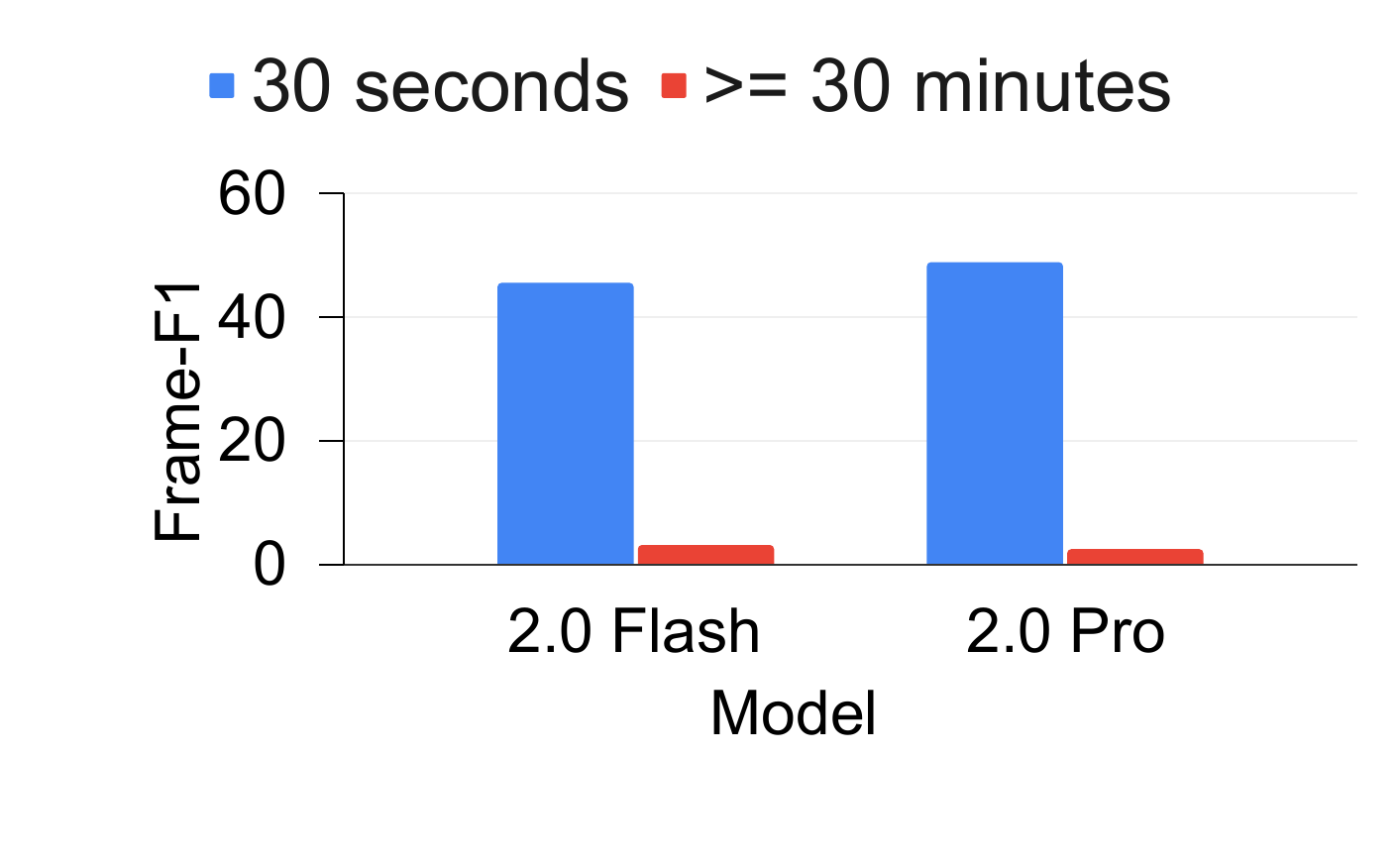}
    \caption{Entity Localization}
    \label{fig7:c}
  \end{subfigure}
  \begin{subfigure}[b]{0.40\linewidth}
    \centering
    \includegraphics[width=\linewidth]{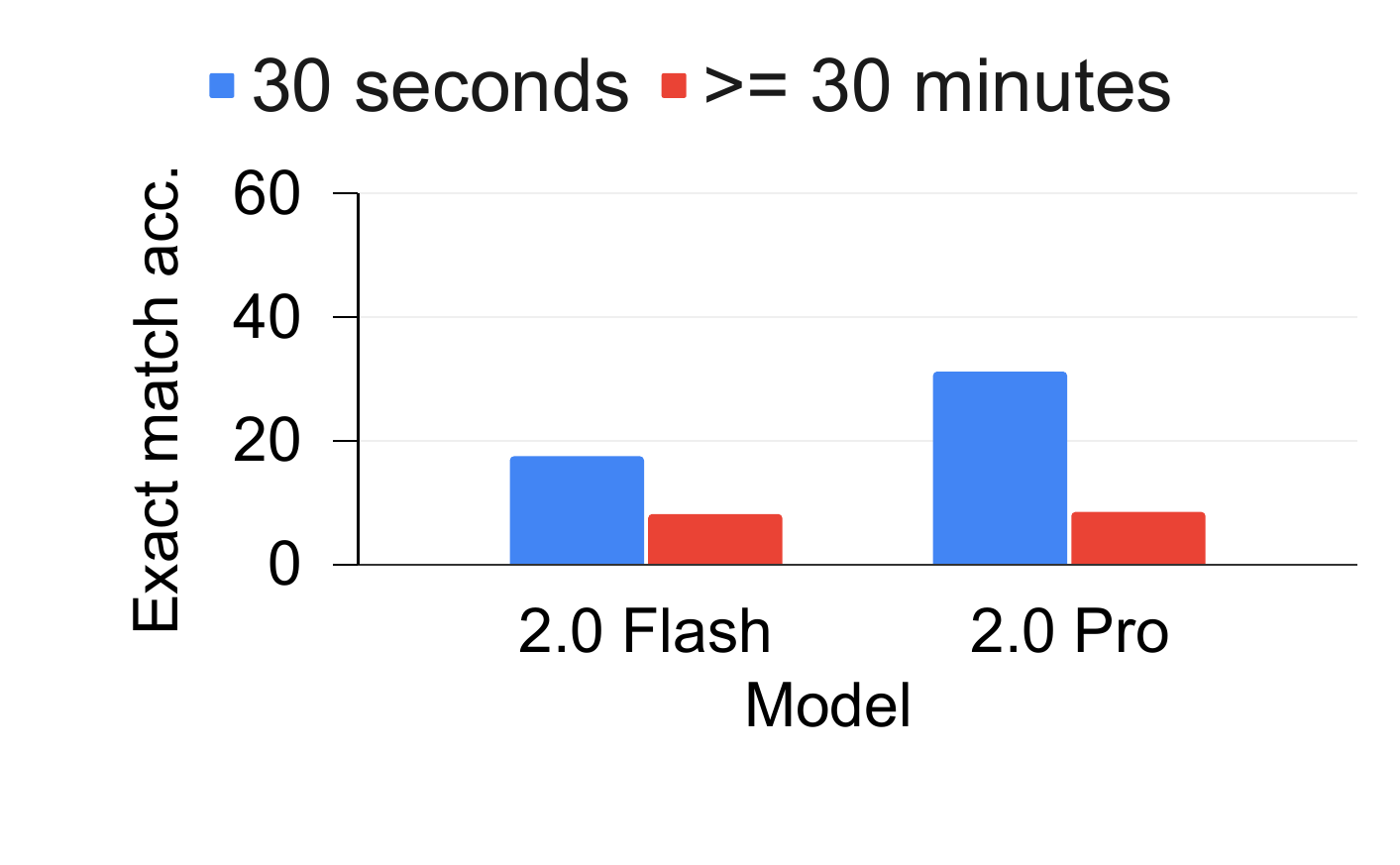}
    \caption{Speaker Number Estimation}
    \label{fig:7:d}
  \end{subfigure}
  \caption{Comparison of long audio and short audio results across Gemini Models}
  \label{fig:short_vs_long_audio_comparison}
\end{figure}

 The results from these analyses are summarized in \autoref{tab:short_audio_evals} and \autoref{fig:short_vs_long_audio_comparison}. As shown in \autoref{fig:short_vs_long_audio_comparison}, Gemini models consistently improve across all tasks as the duration is limited to 30 seconds, with the most visible gains observed in word and NE localization. Meanwhile, \autoref{tab:short_audio_evals} presents a comparison of model performance on \benchmarkmini{} across multiple models, demonstrating that Gemini outperforms all others on every task.

\begin{table}[t]
    \centering
    \resizebox{\columnwidth}{!}{
    \begin{tabular}{l l c c c c c c c  }
        \toprule
        \textbf{Task} & \textbf{Metric} $\uparrow$ & \textbf{G2 Flash} & \textbf{G2 Pro} & \textbf{Q2} &  \textbf{AF2} & \textbf{Phi-4} & \textbf{GPT-4o} \\
        \midrule
        \multirow{1}{*}{Word Localization} 
                           & Word $F_1$   & \textbf{30.22}   &    8.61  & 2.43 & \texttt{-} &  2.73 & \texttt{-} \\
    
         \midrule
        \multirow{1}{*}{NE localization} & Frame-level $F_1$  & 45.49 & \textbf{49.58} & 12.07 & \texttt{-} & 7.63 & \texttt{-}
                          \\
        \midrule
        \multirow{1}{*}{Speaker Number Estimation}  & EMA  & 17.50 & \textbf{31.00} & 7.0 & 6.00 & 15.50  & 14.50 \\
        
        \midrule  
        \multirow{1}{*}{Entire Duration} & EMA  & 5.00/31.00 & 3.50/\textbf{34.50} & 6.5/27.5 & 2.5/20.50 & 3.50/22.00 & \textbf{7.00}/27.00\\
        \midrule
        \multirow{1}{*}{Event Duration}& EMA & 9.45/\textbf{36.22} & 4.72/29.13 & 3.15/18.90 & 1.57/16.54 &  3.9/24.21 & 1.57/18.11\\
        \bottomrule
    \end{tabular}}
    \caption{Performance comparison of audio LMs on \benchmarkmini{} audio tasks ($\le 30$ seconds). \textbf{G2} = Gemini 2.0, \textbf{Q2} = Qwen 2.0, \textbf{AF2} = Audio Flamingo 2. Gemini outperforms all others. Audio Flamingo 2 and GPT-4o refuse to perform any localization task, so we leave them blank. For event duration, we report scores without and with $\pm$2 seconds offset.}
    \label{tab:short_audio_evals}
\end{table}

\paragraph{Are audio LMs long-form zero-shot reasoners? }
In all the results presented so far, we query the model to directly generate the answer. Inspired by test-time compute research \citep{NEURIPS2022_8bb0d291, NEURIPS2022_9d560961}, we explore zero-shot chain-of-thought approaches.  
We append an auxiliary \textit{reasoning prompt} to the original prompt to guide the model in generating reasoning chains or rationales that could lead to better predictions.
We test the following prompts that have been effective in text-only LMs: \textit{``Let's think step by step''} and \textit{``Explain your reasoning before making a prediction''}. We conduct this analysis on speaker number estimation and event duration with Gemini 2.0 Flash and use all long audio examples in \benchmark{}. 

For event duration, we observed accuracies of 6.93 and 6.44 with the reasoning prompts \textit{Let's think step by step''} and \textit{Explain your reasoning before making a prediction''}, respectively—an average improvement of 3\% over the original accuracy of 3.96\%  without any reasoning prompts. For speaker number estimation, the accuracies with both reasoning prompts are 9.55 and 11 respectively, a slight performance boost compared to the original 8\% accuracy. 

We analyzed the individual reasoning steps generated by the models for both prompts, and observed that the model excels at decomposing tasks and producing relevant reasoning traces. However, it sometimes lacks the necessary auditory understanding to accurately execute each sub-task, correctly, which limits its overall performance. For instance, in \autoref{tab:reasoning examples}, we can see that both reasoning prompts enables the model to decompose the task, however the result from each sub-task is not quite correct based on in-correct audio perception. This suggests that while the reasoning traces are good, the model's grounding in the auditory modality remains weak, leading to errors in the final output despite coherent intermediate steps. We conclude that audio language models are not truly zero-shot reasoners; more advanced test-time compute strategies may be needed to better connect high-level reasoning with faithful auditory perception.

\begin{table*}[ht]
\centering
\resizebox{\linewidth}{!}{

\small

\begin{tabular}{ p{3.0cm} p{1.0cm} p{8.5cm} p{5.0cm} }
\toprule 
 \textbf{Question} & \textbf{Ground Truth} & \textbf{ Let's think step by step} & \textbf{Explain your reasoning}   \\
 \midrule
 What is the total time in seconds that the first panelist spoke for throughout the entire podcast? & 1160 
 & \texttt{Here's how we can calculate the total speaking time for the first panelist:1.  **Identify the first panelist:** The first panelist is Dan Mogulof.2.  **Locate the start and end times of Dan's first speaking segment:** Dan starts speaking at [00:02:00] and ends at [00:14:50].3.  **Calculate the duration of Dan's first speaking segment:** 14 minutes and 50 seconds - 2 minutes and 0 seconds = 12 minutes and 50 seconds.4.  **Locate the start and end times of Dan's second speaking segment:** Dan starts speaking at [00:48:30] and ends at [00:56:30].5.  **Calculate the duration of Dan's second speaking segment:** 56 minutes and 30 seconds - 48 minutes and 30 seconds = 8 minutes and 0 seconds.6.  **Convert the durations to seconds:**    *   12 minutes and 50 seconds = (12 * 60) + 50 = 720 + 50 = 770 seconds  *   8 minutes and 0 seconds = (8 * 60) + 0 = 480 seconds7.  **Add the durations together:** 770 seconds + 480 seconds = 1250 seconds.Therefore, the first panelist, Dan Mogulof, spoke for a total of 1250 seconds throughout the entire podcast.}
 & \texttt{Here's how we can estimate the speaking time of the first panelist:1.  **Identify the Speaker:** The first panelist is Dan Mogulof.2.  **Locate the Start Time:** The first panelist begins speaking at 00:01:55.3.  **Locate the End Time:** The first panelist stops speaking at 00:18:00.4.  **Calculate the Duration:** 00:18:00 - 00:01:55 = 00:16:05.5.  **Convert to Seconds:** (16 minutes * 60 seconds/minute) + 5 seconds = 965 seconds.**Prediction:** The first panelist spoke for approximately 965 seconds.} \\

\bottomrule
\end{tabular} }
\caption{Comparison of reasoning prompts and model responses for estimating the duration of a person speaking in a panel discussion. While the model effectively decomposes the task and generates relevant reasoning, it often lacks the auditory understanding needed to execute sub-tasks accurately.}
\label{tab:reasoning examples}
\end{table*}

\paragraph{Robustness of audio LMs on long-form audio}
Recent work has suggested that certain audio LMs rely only on text prompts, rather than jointly attending to text and audio input. 
\citet{sakshi2024mmaumassivemultitaskaudio} report that GAMA \citep{ghosh-etal-2024-gama}, Qwen2-Instruct \citep{Qwen2-Audio} and Gemini Pro \citep{geminiteam2024gemini15unlockingmultimodal} are more robust to noisy audio and are usually more attentive to audio content compared to other models like SALMONN \citep{tang2024salmonn}
. However, these experiments were done on short audio samples (up to 30 seconds).  

\begin{wrapfigure}{r}{0.4\textwidth}
    \centering
    \includegraphics[width=1\linewidth]{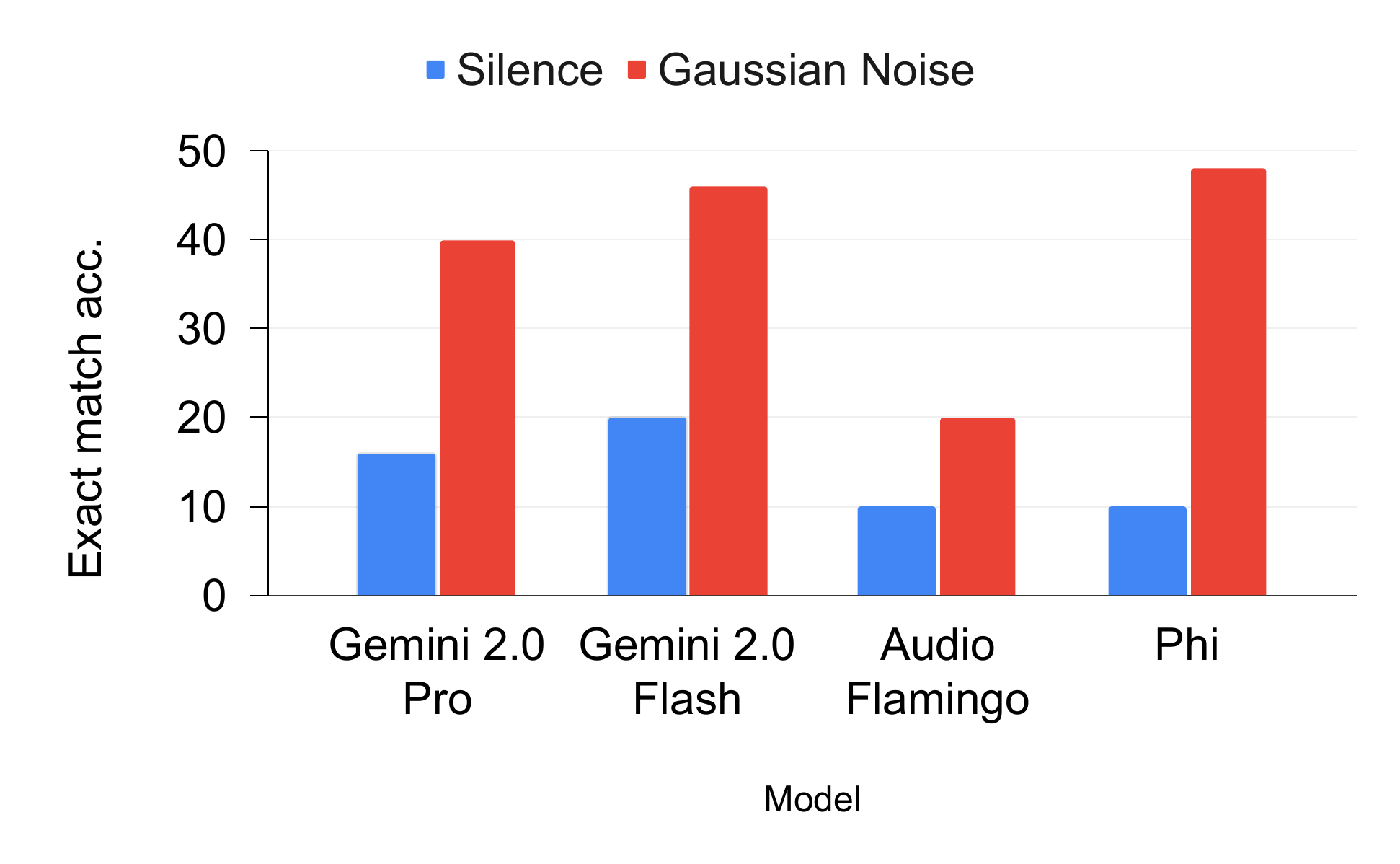}
    \caption{Performance comparison when the original audio input is replaced with silence or Gaussian noise. Since the entire input is noisy, the ground truth label is 0}
    \vspace*{-10pt}
    \label{fig:robustness_snv}
\end{wrapfigure}

We follow \citet{sakshi2024mmaumassivemultitaskaudio} 
in this analysis and start by comparing the original model's prediction for speaker number estimation with its prediction when the audio input is replaced with random Gaussian noise. In a different setting, we also replaced the original audio input with silence. All of our experiments are performed on Gemini, Phi-4-Mini and Audio Flamingo, since they support longer durations than other models. We generate 5 minute noisy audio samples for Audio Flamingo and Phi-4-Mini, while we generate one hour long noisy audios samples for Gemini.
In contrast to previous work on short audio \citep{sakshi2024mmaumassivemultitaskaudio}, which finds that audio language models (LMs) are robust to noisy short audio samples, our findings reveal a different trend for long noisy inputs. Specifically, our analysis shows in \autoref{fig:robustness_snv} that the models are not robust to noisy inputs, and they are particularly less robust to silence than to Gaussian noise.

Next, we investigated how the positioning of noise affects the robustness of an audio LM. Unlike our previous analysis, where the entire input was replaced with Gaussian noise or silence, we now introduce 30-second audio clips from \benchmarkmini{} into 60-minute noisy recordings for Gemini and 5-minute noisy recordings for Phi and Audio Flamingo, placing the clips at various positions. Our goal is to measure the models ability to model's ability to disregard background noise and focus on meaningful content. We conducted this experiment for speaker number estimation, varying the placement of the clean audio clip based on the model's maximum input duration. 

\begin{figure}[h]
  \centering
  \begin{subfigure}[b]{0.48\linewidth}
    \centering
    \includegraphics[width=\linewidth]{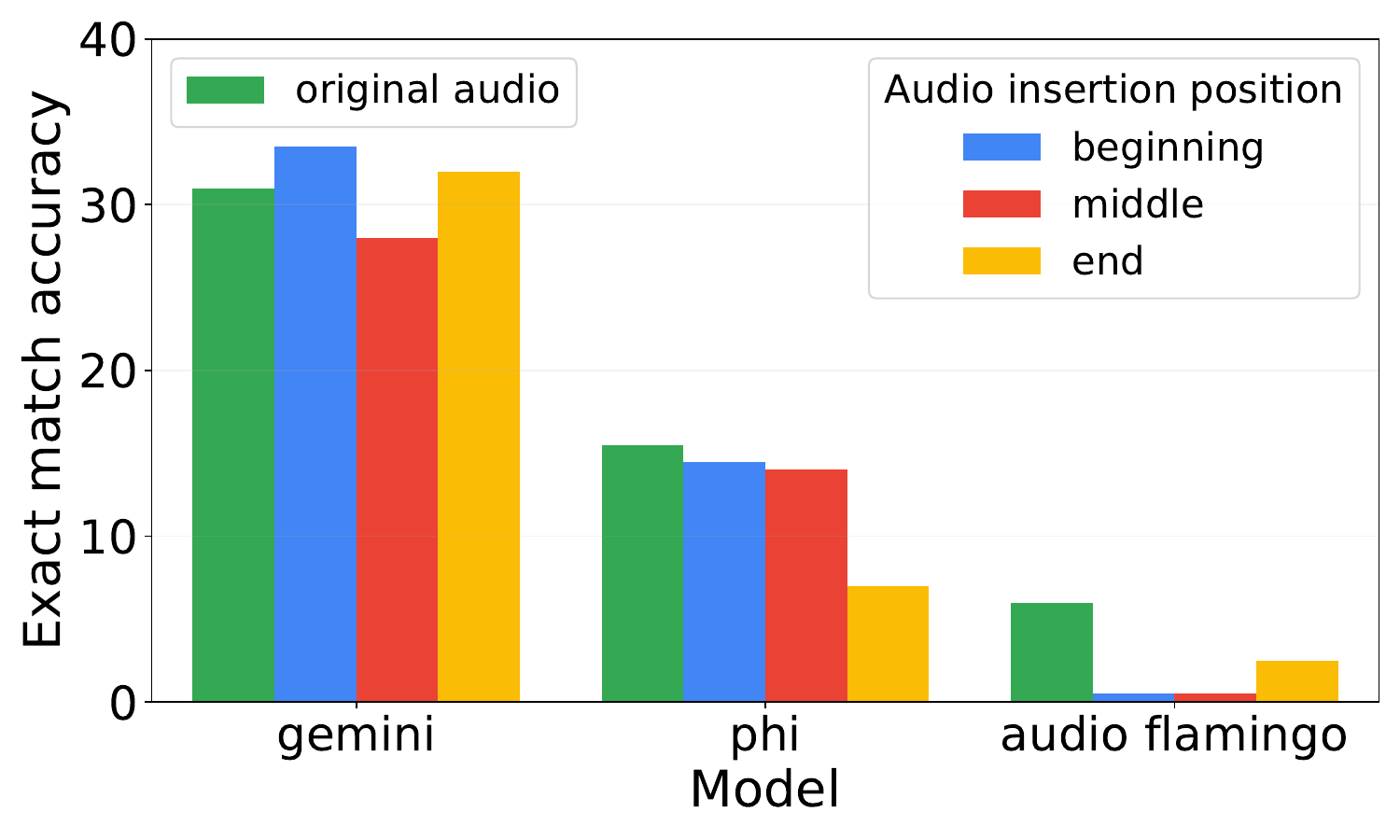}
    \caption{Silence}
    \label{fig:roba}
  \end{subfigure}
  \hfill
  \begin{subfigure}[b]{0.48\linewidth}
    \centering
    \includegraphics[width=\linewidth]{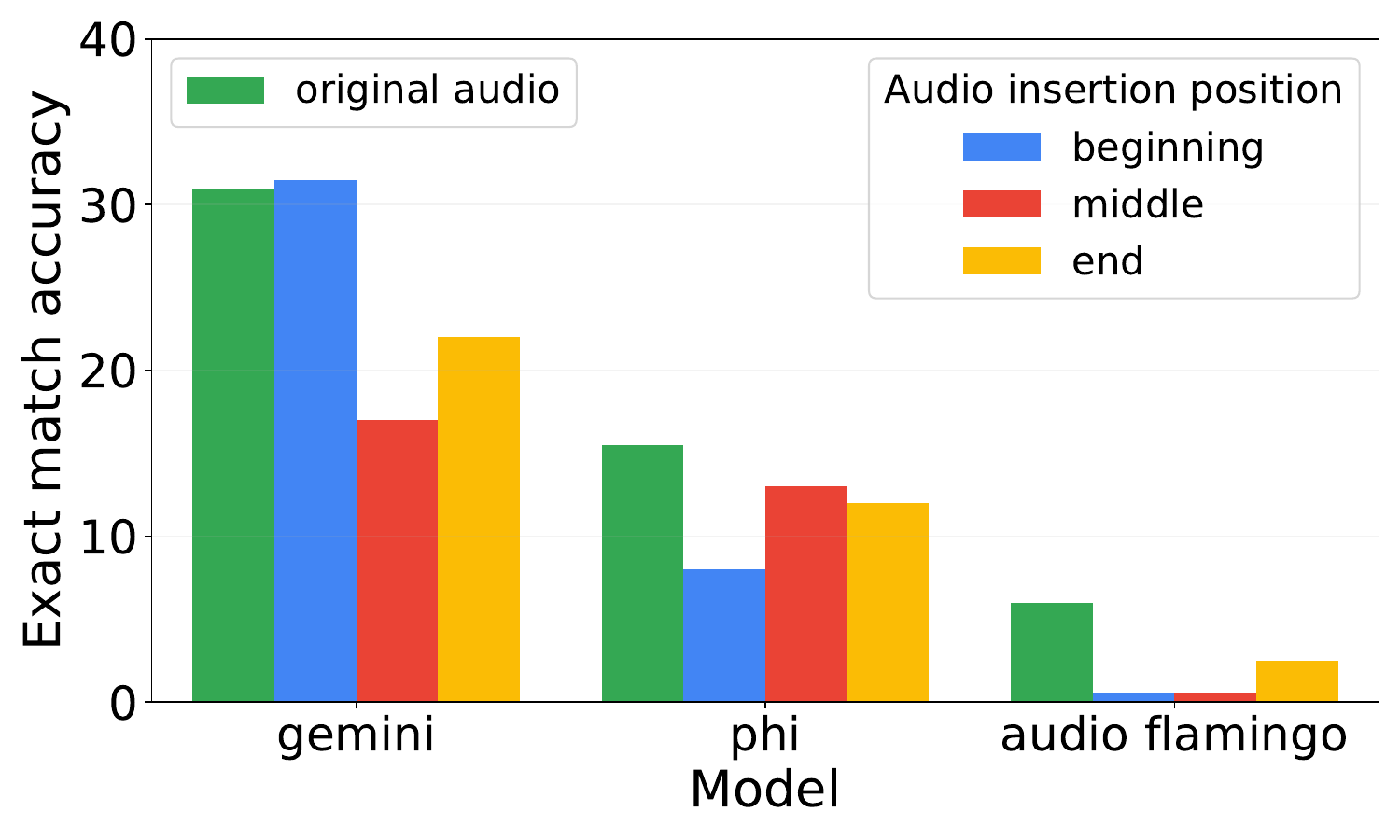}
    \caption{Gaussian}
    \label{fig:robb}
  \end{subfigure}
  \caption{Placing a 30-second clean audio clip at different points within a long, noisy audio input impacts speaker number estimation performance. Proprietary models like Gemini perform better when the clean clip is positioned at the beginning or end of the noisy audio.}
  \label{fig:robustness_position}
\end{figure}

In \autoref{fig:robustness_position}, we compare the performance of Gemini, Phi, and Audio Flamingo. For Gemini, we observe that performance degrades when the relevant 30-second audio clip is placed in the middle of the input, suggesting that the model struggles to effectively use information at the center in long input contexts. This is consistent with previous findings on text-based LMs, where performance tends to be highest when relevant information appears at the beginning or end of the input and significantly drops when it is located in the middle, even for models designed to handle long contexts \citep{liu-etal-2024-lost}. Additionally, we find that performance degradation is more pronounced when the surrounding audio contains Gaussian noise compared to silence. The trends for the 5-minute audio LMs differ. Phi performs better when the 30-second clip is placed in the middle of the noise, while Audio Flamingo shows consistent degradation, with a slight improvement at the end. We hypothesize that Audio Flamingo generally struggles to distinguish noise from actual audio across the entire 5-minute clip, as seen in \autoref{fig:robustness_snv}.

%% file: 06.tex
\section{Related work }
\paragraph{Audio Benchmarking}
Audio benchmarks can be broadly classified into two main categories based on their scope and purpose: \textit{Task-Specific Benchmarks} focus on evaluating models' performance on particular audio tasks. Examples include text-audio retrieval~\citep{Koepkeetal2021}, compositional audio reasoning~\citep{Ghoshetal2023}, automatic speech recognition~\citep{Panayotovetal2015, shi24g_interspeech}, audio captioning~\citep{Drossosetal2019,Kimetal2019}, and emotion recognition~\citep{Livingstoneetal2018}. Several benchmarks combine such tasks into a collection such as SUPERB~\citep{Yangetal2021},  HEAR~\citep{Turianetal2022}, among others. Our work falls under the umbrella of \textit{instruction following benchmarks} which assess model capabilities to understand audio signals and follow instructions in a conversation format. Dynamic-SUPERB was one of the first benchmarks of this kind~\citep{Huangetal2023}, followed by AIR-Bench~\citep{yang-etal-2024-air} and MMAU~\citep{sakshi2024mmaumassivemultitaskaudio}. However, almost all of these benchmarks contain  samples with a maximum duration of 30 seconds. Most closely related to our work is \citet{ghosh2025audioflamingo2audiolanguage}, who release LongAudioBench containing tasks with samples with a maximum duration of 5 minutes. Our work focuses on much longer long audio samples which can be up to 2 hours long.

\paragraph{Long Context Benchmarks}
As the context length for text-based LMs has increased, new benchmarks have emerged to test how well the models handle long-form text. These include tasks like information retrieval (Needle-in-a-Haystack; \citealp{nelson2024needle}) and synthetic long-range reasoning (Long Range Arena;~\citealp{tay2020long}). Newer benchmarks such as LongBench~\citep{Baietal2023} evaluate models on tasks over inputs with up to 128K tokens. While these efforts have advanced long-context evaluation in text, similar benchmarks are largely missing for audio. Our work fills this gap by introducing a benchmark for evaluating models on extremely long-form audio understanding.

\section{Conclusion }
In this paper, we introduce Brutally Long Audio Bench (BLAB), a challenging benchmark designed to evaluate audio language models on localization, duration estimation, emotion, and counting tasks. \benchmark{} is the first benchmark to assess audio LMs on long-form audio, with durations ranging from over 15 minutes to up to 2 hours. 
Our evaluation of six open-weight and proprietary audio LMs reveals that these models struggle substantially with long-form speech, with performance deteriorating as audio duration increases. Additionally, we find that audio LMs perform poorly on both temporal and counting tasks and struggle to process non-phonemic information in audio.  
Contrary to existing findings on short-form audio, our analysis suggests that audio LMs are not particularly robust when handling long-form speech. These models tend to rely more on prompts than on actual audio content, making them susceptible to distractions from noisy audio, such as Gaussian noise and silence.  We provide a detailed report of our data curation and evaluation framework. Overall our findings underscore the need for more approaches to developing long-context  multimodal language models with strong long-form audio understanding capabilities. 

\section*{Limitations}
Audio data for \benchmark{} is sourced from various real-world recordings, including interviews, podcasts, and political speeches.
Our annotation framework leverages human-AI collaboration. However, we observe that overlapping speech is common, which can impact the accuracy of automatic annotation tools like WhisperX and Pyannote \citep{bain2022whisperx, Bredin23}, as they may not be robust in handling such cases. To address this, our framework includes human verification steps to improve annotation quality.

%% file: appendix.tex
\newpage
\section{Appendix}

\subsection{Metrics }
\label{sec:metrics}
Given a model output, we post-process it to match the expected ground-truth format. For localization tasks, we use the \texttt{json\_repair} library.\footnote{https://pypi.org/project/json-repair/} For the remaining tasks with numeric outputs, we use regular expressions to extract relevant numerical values. The regular expression are designed to identify and format various numeric formats, including integers and numbers expressed with units (e.g., "35 seconds" or 00:23:00). 
\paragraph{Localization Tasks} For these tasks, the models are expected to generate JSON outputs. 
Word localization is typically evaluated using metrics that compare start and end timestamp differences for matching words in the audio transcript, detecting delays in word onset or offset \citet{Sainath2020EmittingWT}. For ease of evaluation, we report $F_1$ scores on the number of correctly aligned words.

For NE and advertisement localization, which are span-localization tasks, we use Frame-level $F_1$ as mentioned in \citet{shon-etal-2023-slue}. This metric is derived from question-answering evaluation frameworks~\citep{chuang20b_interspeech}, and measures the overlap between the predicted and the ground truth answer spans.

\paragraph{Counting and Duration and Emotion tasks}
These tasks require numeric answers and are evaluated using exact match accuracy (EMA).

\subsection{Dataset Statistics of \benchmarkmini{}}
\benchmarkmini{} is a subset of \benchmark that contains audio samples less than or equal to 30 seconds of audio on average.
\begin{table}[h!]
    \centering
    \resizebox{0.8\columnwidth}{!}{
    \begin{tabular}{c c c c }
        \toprule
        \textbf{Task} & \textbf{TD (mins.) } & \textbf{AD (seconds.)} & \textbf{Number of questions} \\
        \midrule
        \multirow{1}{*}{Word Localization}    & 89.00   & 30.00   &    178   \\
    
         \midrule
        \multirow{1}{*}{NE localization} & 36.00   & 30.00   &    107   \\
        \midrule
        \multirow{1}{*}{Speaker Number Estimation}  & 99.17   & 30.00   & 200  \\
        \midrule  
        \multirow{1}{*}{Entire Duration} & 58.78  & 17.64   &    200 \\
        \midrule
        \multirow{1}{*}{Event Duration}& 63.55   & 29.79   &    128   \\
        \midrule
      \textbf{Total} & \textbf{346.5}   &  &   \textbf{813}  \\    
        \bottomrule
    \end{tabular}}
   \caption{Statistics and examples of questions for each task in \benchmarkmini{}. TD and AD represent total and average duration, respectively.}
    \label{tab:short_audio_data_statistics}
\end{table}



\newpage
\subsection{Prompt formatting}
\begin{center}
\begin{longtable}{p{4cm} p{9cm}}
\toprule
 \multicolumn{1}{c}{\textbf{Task}} & \multicolumn{1}{c}{\textbf{Prompt Format}}\\ 
 \midrule
\endfirsthead

\multicolumn{2}{c}%
{{\bfseries \tablename\ \thetable{} -- continued from previous page}} \\
\hline \multicolumn{1}{c}{\textbf{Task}} & \multicolumn{1}{c}{\textbf{Prompt Format}}\\ \hline 
\endhead

\hline \multicolumn{2}{|r|}{{Continued on next page}} \\ \hline
\endfoot

\hline \hline
\endlastfoot
           Word Localization  & 
           
            \texttt{Align the provided audio file with each word in its transcript. For each word spoken, predict the start and end timestamps in seconds and milliseconds.} \newline 
           
          \texttt{Format the response as:}
          
        \texttt{{\{\{'word': '<word>', 'start': <start timestamp of the word>, 'end': <end timestamp of the word>\}\}\}}}

           \\
           \midrule
           NE Localization  & \texttt{Your task is to analyze an audio file and detect all \textbf{movie} named entities present within it. \textbf{Movie} entity refers to the name of \textbf{feature films or animated movies}. For each detected \textbf{movie} entity, provide the \textbf{start} and \textbf{end} timestamps (in seconds and milliseconds) that indicate the location and duration of the entity in the audio. } \newline 
           
           \texttt{**Expected Output Format**} \newline  
        \texttt{Your response should be formatted as a list of dictionaries:}
        
        \texttt{{\{\{
        [
            {'entity': '<movie name>', 'category': 'movie', 'start': <start timestamp>, 'end': <end timestamp>},
            {'entity': '<movie name>', 'category': 'movie', 'start': <start timestamp>, 'end': <end timestamp>}
        ]
        \}\}\}}}
           
           \\
           \midrule
            Advertisement Localization  &  \texttt{Given the audio file, your task is to detect the presence of promotions and advertisements within the audio. You are required to predict the start and end timestamps (in seconds and milliseconds) for the entire span for all promotion and advertisement segments you have identified.} \newline

            \texttt{ Format the response as:}\newline  
    \texttt{{\{\{ {{"advertisement": <advertisement text>, "start": <start timestamp of the advertisement>, "end": <end timestamp of the advertisement>}}\}\}\}}}

            \\
            \midrule
           Speaker Number Estimation  &  \texttt{What's the number of unique voices detected in this sound file?} \newline
            
           \texttt{Provide only the numeric value without any explanation.}\\
           \midrule
            Valence \& Arousal Ranking  & \texttt{How would you assess the valence of the speaker in the fourth apology clip within this audio file?} \newline
            
            \texttt{Listen to the audio and select one option from the provided choices that best matches the answer. Return only that option.} 
            
             \texttt{Options: \newline 
        (A) Very Pleasant \newline 
        (B) Pleasant \newline 
        (C) Neutral \newline 
        (D) Very Unpleasant \newline 
         (D) Unpleasant \newline 
        }

            \\
            \midrule
           Emotion Reasoning &  \texttt{Throughout the course of this debate, which speaker demonstrates the most consistent emotional intensity, particularly in their tone and delivery?} \newline  
           
           \texttt{Listen to the audio and select one option from the provided choices that best matches the answer. Return only that option.} 
        \texttt{Options: \newline 
        (A) The Tory Party leader \newline 
        (B) The Labour Party leader \newline 
        (C) Both speakers exhibit similar levels \newline 
        (D)  It is difficult to determine \newline 
        }

           \\
           \midrule
            Entire Duration  &  \texttt{What is the total duration of this audio file in seconds only?} \newline
            
            \texttt{Provide only the numeric value without any explanation.}\\
            \midrule
           Event Duration & \texttt{A harmonica is being played fifteen seconds towards the end of this audio. How long in seconds was this instrument played for ?}
            \texttt{Provide only the numeric value as an integer without any explanation. Do not use the MM:SS format.}
           \\
            \bottomrule
  
     \caption{Exact prompt formats used for evaluating each task in \benchmark{}.} 
    \label{tab:prompt_format}
   
\end{longtable}

\end{center}